\documentclass[conference,letterpaper]{IEEEtran}

\usepackage{algpseudocode,algorithm,amsmath,amsfonts,amssymb,amsthm,bm,booktabs,graphicx,lipsum,subcaption,tikz,url,xcolor,xifthen}
\usepackage[pass]{geometry}
\let\labelindent\relax
\usepackage{enumitem}
\usepackage{verbatim} 
\usepackage{hyperref} 
\usepackage{cleveref} 

\usepackage{microtype}  

\creflabelformat{equation}{#2#1#3}
\usetikzlibrary{chains,fit,shapes,patterns}

\graphicspath{ {../img/} }

\theoremstyle{definition}

\newcommand{\xopt}{\bm{x}_\text{opt}}

\newcommand{\y}[2][]{
    \ifthenelse{\isempty{#1}}
    { y^{#2} }      
    { y^{#2}_{#1} } 
}
\newcommand{\avgy}[1]{
    \overline{y}^{#1}
}

\renewcommand{\theenumii}{\arabic{enumii})}

\begin{document}

\title{Evolving the Structure of Evolution Strategies} 

\author{
    \IEEEauthorblockN{Sander van Rijn, Hao Wang, Matthijs van Leeuwen, Thomas B\"{a}ck}
    \IEEEauthorblockA{
        Natural Computing Group\\
        LIACS, Leiden University\\
        Niels Bohrweg 1, 2333 CA\\
        Leiden, The Netherlands\\
  		Email: \{s.j.van.rijn, h.wang, m.van.leeuwen, t.h.w.baeck\}@liacs.leidenuniv.nl
    }
}

\IEEEoverridecommandlockouts
\IEEEpubid{\makebox[\columnwidth]{978-1-4799-7492-4/15/\$31.00~
\copyright2016
IEEE \hfill} \hspace{\columnsep}\makebox[\columnwidth]{ }}

\maketitle

\begin{abstract}
    Various variants of the well known Covariance Matrix Adaptation Evolution Strategy (CMA-ES) have been proposed recently, which improve the empirical performance of the original algorithm by structural modifications. However, in practice it is often unclear which variation is best suited to the specific optimization problem at hand. As one approach to tackle this issue, algorithmic mechanisms attached to CMA-ES variants are considered and extracted as functional \emph{modules}, allowing for combinations of them. This leads to a configuration space over ES structures, which enables the exploration of algorithm structures and paves the way toward novel algorithm generation. Specifically, eleven modules are incorporated in this framework with two or three alternative configurations for each module, resulting in $4\,608$ algorithms. A self-adaptive Genetic Algorithm (GA) is used to efficiently evolve effective ES-structures for given classes of optimization problems, outperforming any classical CMA-ES variants from literature. The proposed approach is evaluated on noiseless functions from BBOB suite. Furthermore, such an observation is again confirmed on different function groups and dimensionality, indicating the feasibility of ES configuration on real-world problem classes.
\end{abstract}

\section{Introduction}

Evolutionary Algorithms (EAs) such as Genetic Algorithms (GAs) \cite{holland_adaptation_1975} and Evolution Strategies (ESs) \cite{rechenberg_evolutionsstrategie_1973} have been studied for decades, leading to the many variants proposed in the literature \cite{back2013contemporary}. The performance of all these algorithms depends not only on the specific optimization task, but also on appropriately tuning the algorithm's parameters such as population size and mutation rate. As a result, the ideas of online tuning \cite{grefenstette_optimization_1986} and automated parameter optimization have been proposed \cite{back96}. These were later reinforced by the realization that an increase in performance for some problem instances will necessarily cause a decrease in performance for some other instances. This implies that the potential performance improvement that can be gained from optimizing parameters is always limited by the algorithm that is chosen. To limit the bias imposed by the choice of a specific (evolutionary) algorithm, researchers have proposed to instead \emph{evolve the structure of an EA itself} (e.g., \cite{kantschik_meta-evolution_1999,kantschik_empirical_1999,spector_genetic_2002}).

The class of optimizers derived from the Covariance Matrix Adaptation Evolution Strategy (CMA-ES) \cite{hansen1996adapting} are the preferred optimization method for many real-valued black-box problems, and are therefore also the scope of this paper. As discussed in more detail shortly, many variations of the CMA-ES have been separately introduced and discussed in the literature. However, only few combinations of these variations have been empirically tested and compared. As a result, in practice it is often unclear which variation is best suited to the optimization task at hand. This leads to the three main questions in this work:
\begin{enumerate}
	\item Can we define a modular and extensible CMA-ES framework that allows to unify the many variations that have been introduced in the literature?
	\item Within the framework defining a large number of CMA-ES variations, how to determine an efficient ES structure, given limited function evaluation budget?
	\item Are there any novel variations, i.e., combinations of methods that have been proposed in the literature, that outperform the known variants?
\end{enumerate}

\textbf{Approach and contributions}
Firstly, a modular and extensible CMA-ES framework is proposed based on the original CMA-ES \cite{hansen1996adapting}, for adapting the structure of Evolution Strategies in particular. A number of independent, functional \textit{modules} are extracted from various existing ES variants. By allowing each of these modules to be activated independently, a set of $4\,608$ so-called \emph{ES-structures} can be instantiated, many of which have never been considered before. A detailed overview of the framework and all selected modules is given in \Cref{sec:cmaes-frame}.

Secondly, a metaheuristic is proposed to search in the configuration space containing ES-structures.
In \Cref{sec:ga} a simple yet effective genetic encoding scheme is used, facilitating a mutation-only and self-adaptive GA optimization. Although it is possible to deploy an exhaustive search (brutal force) on such a combinatorial configuration space, the time complexity of exhaustive search would grow polynomially with increasing number of ES modules and makes it computationally infeasible in practice. Additionally, \Cref{sub:eval} investigates how we can robustly evaluate evolution strategies. The proposed approach combines a stable fitness measure---based on \textit{Fixed Cost Error} (FCE) and \textit{Estimated Running Time} (ERT)---with a statistical estimation of the number of runs required to reliably compare different strategies.

Finally, the third contribution consists of an extensive empirical evaluation to investigate the potential of  the CMA-ES configuration framework and metaheuristic used for the configuration. The experiments described in \Cref{sec:exp}, using the noiseless optimization functions from BBOB \cite{hansen:bbob}, show that the meta-GA consistently converges quickly and produces results that are on par with the best possible configurations found by brute force search, needing only 5\% of the evaluations. Furthermore,  the effectiveness of various modules is shown on each function group and dimensionality, indicating that searching for a suitable ES structure is a better solution than always resorting to a single `default' configuration.

\section{Related Work}
\label{sec:prevwork}

The idea of evolving the structure of Evolutionary Algorithms (EAs) is first motivated in~\cite{koza_genetic_1992}, using the Genetic Programming (GP) technique. Multiple attempts have been made for this purpose, e.g. meta-evolution on graph-based programs~\cite{kantschik_meta-evolution_1999,kantschik_empirical_1999}. In those studies, varying numbers of meta-levels are used, each responsible for evolving the program used in the level below and potentially itself. The \textit{Push} language~\cite{spector_genetic_2002} was later dedicated to enable autoconstructive evolution whereby a population of programs is expected to produce other programs.

Later approaches focused more on applying evolution within the existing structure of an EA. In \cite{tavares_evolution_2004} GP is used to create new mutation, recombination and selection operators, from which a standard EA structure is constructed. Oltean \textit{et al.} use existing operators as the genes to create new structures in \cite{oltean_evolving_2003,oltean_evolving_2003-1,oltean_evolving_2005}, while tree-based structures of multiple EAs with different parameters are evolved in \cite{martin_evolving_2013} by Martin \textit{et al}.

The Grammatical Evolution (GE) approach by Louren{\c c}o \textit{et al.} \cite{lourenco_evolving_2012} is highly related to this work. The structure of an EA is represented as a context free grammar with parameters and operators as terminals. Treating an EA as a collection of atomic operators in a fixed framework allows arbitrary combinations, similar to our proposed approach.

\section{CMA-ES Framework}
\label{sec:cmaes-frame}

This section introduces the modular and extensible CMA-ES framework. Short summaries of all considered modules are first given in \Cref{sub:es}, after which \Cref{sub:framework} introduces the modular framework. Implementation details are described in \Cref{sub:imp}.

\subsection{ES Variations}
\label{sub:es}

Eleven possible modules are considered in total, nine of which have two available options, and the remaining two have three options. This results in $2^9 \cdot 3^2 = \mathbf{4\,608}$ different ES-structures. For each module, a brief description is given below.

\begin{enumerate}
    \item
    \textbf{Active Update:}
    The update of covariance matrix $C$ is normally only done by taking the most successful mutations into account. The Active Update, introduced by Jastrebski \textit{et al.} \cite{jastrebski2006active}, adapts the covariance matrix using the negative factor based on the least successful individuals, in addition to the standard covariance matrix update.

    \item
    \textbf{Elitism:}
    Both $(\mu, \lambda)$- and  $(\mu{+}\lambda)$-strategy are proposed together in evolutionary algorithms. In this work, elitism is considered as an alternative.

    \item
    \textbf{Mirrored Sampling:}
    A technique to ensure more evenly spaced sampling of the search space is Mirrored Sampling by Brockhoff \textit{et al.} \cite{brockhoff2010mirrored}. Half of the mutation vectors are still sampled from the normal distribution, but every other mutation vector is the mirror image of the previous random vector.

    \item
    \textbf{Orthogonal Sampling:}
    A later addition to Mirrored Sampling was Orthogonal Sampling by Wang \textit{et al.} \cite{wang2014mirrored}. The desired number of samples is first drawn from the normal distribution. The Gram-Schmidt process is then used to orthonormalize the set of vectors.

    \item
    \textbf{Sequential Selection:}
    Without parallel execution, all $\lambda$ individuals are first evaluated in order, and then selection is applied. The sequential selection method proposed by Brockhoff \textit{et al.} \cite{brockhoff2010mirrored} immediately compares the function value of each newly evaluated individual to the best found so far, and does not evaluate any additional offspring individuals when an improvement has been found.

    \item
    \textbf{Threshold Convergence:}
    Becoming stuck in a local optimum is a common problem when using an ES. Piad \textit{et al.} propose Threshold Convergence \cite{piad2015evolution} as a method of forcing the evolution to stay in an exploratory phase for longer, by requiring mutation vectors to reach a length threshold. This threshold then decreases after every generation to slowly transition into local search.

    \item
    \textbf{Two-Point Step-Size Adaptation (TPA):}
    The step size $\sigma$ of the CMA-ES is adapted after every generation according to the evolution path, which incorporates the latest successful individuals. Hansen \textit{et al.} proposed TPA \cite{hansen2008tpa}, which reserves two individuals from the $\lambda$ offspring. These are used to evaluate two mutations after selection and recombination has taken place: one with a longer, the other with a shorter version of the weighted average mutation vector belonging to the $
    \mu$ selected individuals. Which of these two results in a lesser function value, determines whether the step-size should increase or decrease.

    \item
    \textbf{Pairwise Selection:}
    Mirrored Sampling can cause a bias in the length of mutation vectors, as two mirrored vectors will (partially) cancel each other out in recombination. Pairwise Selection was introduced by Auger \textit{et al.} \cite{auger2011mirrored} to prevent this. In this paper, the best offspring is first selected from each mirrored pair. The regular selection operator is then applied to all offspring that were selected in this previous step.

    \item
    \textbf{Recombination Weights:}
    In the CMA-ES, recombination is performed with the following weight vector: $$w_i = \log(\mu+\frac{1}{2}) - \frac{\log(i)}{\sum_j w_j}$$ for $i=1,\ldots,\mu$. Alternative weights are the arithmetic mean $w_i = \frac{1}{\mu}$.

    \item
    \textbf{Quasi-Gaussian Sampling:}
    Samples are not necessarily uniformly drawn from the normal distribution. Alternatively, the vectors can be drawn from a quasi-random uniform sequence, which are then transformed to a Gaussian distribution, as proposed in \cite{auger2006algorithms}. As quasi-random sequences, the \textit{Sobol} and \textit{Halton} sequences can be used.

    \item
    \textbf{Increasing Population Size (IPOP):}
    Restarting an ES can be done when no improvements have been found in recent generations. Auger \textit{et al.} proposed an increasing population scheme IPOP \cite{auger2005restart} to use the remaining function evaluations more effectively after a restart. Later, Hansen \textit{et al.} introduced the bi-population (BIPOP) \cite{hansen2009benchmarking} variation which alternates between a larger and smaller population size.

\end{enumerate}

Note that although pairwise selection and orthogonal sampling were introduced in combination with mirrored sampling, each can be activated independently in this framework.

\subsection{ES Framework}
\label{sub:framework}

To easily allow the combination of all ES-variations listed in \Cref{sub:es}, we introduce a modular framework based on the CMA-ES (see \Cref{alg:es}). It is designed such that a module can be activated by replacing a function or passing an additional variable. Any endogenous variables of the CMA-ES and its variations are abstracted into a single global \textit{params} object (lines 9, 14) that is accessible from all other functions.

Default values are available in literature for all relevant parameters, whether belonging to the standard CMA-ES or to any particular variant. These defaults are only used when no other values are specified at initialization (line 6).

\begin{algorithm}
    \caption{Modular CMA-ES Framework}
    \label{alg:es}
    \algrenewcommand\algorithmicindent{1.0em}
    \begin{algorithmic}[1]
        \State options $\gets$ which modules are active
        \State \textit{// Local restart loop}
        \While {not terminate}
            \State $t \gets 0$
            \State $\bar{x} \gets$ randomly generated individual
            \State SetParameters(init-params)
            \State \textit{// ES execution loop}
            \While {not terminate local}
                \State params $\gets$ Initialize(init-params)
                \State $\vec{x} \gets $ Mutate($\bar{x}$, options) \hspace{30pt} \textit{// Sampler, Threshold}
                \State $\vec{f} \gets $ Evaluate($\vec{x}$, options) \hspace{29pt} \textit{// Sequential}
                \State $P^{(t+1)} \gets $ Select($\vec{x}$, $\vec{f}$, options) \hspace{10pt} \textit{// Elitism, Pairwise}
                \State $\bar{x} \gets$ Recombine($P^{(t+1)}$, options) \textit{// Weights}
                \State UpdateParams(params, options) \hspace{4pt} \textit{// Active, TPA}
                \State $t \gets t+1$
            \EndWhile
            \State AdaptParams(init-params) \hspace{38pt} \textit{// (B)IPOP}
        \EndWhile
    \end{algorithmic}
\end{algorithm}

The variable functions are the mutation (line 10), selection (line 12), recombination (lines 13) and parameter update (line 14). Here, the variability is shown by the added \textit{options} argument in all function calls. The sampler is a special case that merges three variations: Quasi-Gaussian sampling replaces the regular Gaussian sampling that is used as \textit{base-sampler}. Mirrored sampling and Orthogonal sampling are added on top of the base-sampler instead.

Sequential selection (line 11) is performed in the abstracted \textit{Evaluate} function that otherwise acts as a simple call-through to the evaluation function. The local restart criteria checked in line 8 are all those from \cite{auger2005restart,hansen2009benchmarking}, and population size adaptations for (B)IPOP are performed in line 17.

Each module can be activated independently of all others. Furthermore, new variations on existing modules can easily be created, and new modules can be added without rewriting the entire algorithm.

\subsection{Implementation Details}
\label{sub:imp}

Our aim is to allow any possible combination of the modules listed in \Cref{sub:es} such that the resulting ES will run with minimal need of checking dependencies between variants, and without causing runtime errors. Two of the considered modules have to be adapted before they can be used in this way, because they were proposed for different evolution strategies.

\begin{itemize}
    \item 
    \textbf{Sequential Selection:}
    Originally intended to be used in $(1, \lambda)$-strategies, a delay in the cut-off is introduced to ensure evaluation of at least $\mu$ individuals when $\mu > 1$. This represents a more robust solution than accepting $< \mu$ individuals and adapting all following calculations.

    \item
    \textbf{Threshold Convergence:}
    In the original paper by Piad \textit{et al.} \cite{piad2015evolution}, Threshold Convergence is used in a regular $(\mu, \lambda)$-ES. The threshold is applied to the mutation vector after it has been scaled by the step size $\sigma$. If this method is equally applied in our framework, the benefit of the CMA-ES is lost for small mutations, because the covariance matrix $C$ scales mutations in different directions differently. Instead, the threshold is applied to the randomly sampled vector, before it is used in any further calculations. This forces the mutation vector to have a minimal length, without losing benefits of the covariance matrix $C$.

\end{itemize}

Furthermore, selection modules can cause problems when they require more than $\mu$ individuals for the selection process. Descriptions of the encountered issues and our solutions are given below.

\begin{itemize}
    \item 
    \textbf{Pairwise Selection and Sequential Selection:}
    For pairwise selection to return $\mu$ individuals, a selection must be made from at least $2\mu$ individuals. This causes a problem when sequential selection is allowed to stop the generation after $\mu$ individuals. To solve this, the cut-off point for sequential selection is artificially increased to $2\mu$. If $\lambda < 2\mu$, $\lambda$ is also increased to $2\mu$.

    \item
    \textbf{Pairwise Selection and TPA:}
    TPA reserves two individuals from the $\lambda$ offspring, preventing them from being used for selection and recombination. This leaves the ES with $\lambda_{\text{eff}} = \lambda-2$ individuals. When pairwise selection is used and $\lambda = 2\mu$, we are one pair short of being able to select $\mu$ individuals. In this case, $\mu$ is set to $\lambda_{\text{eff}}/2$.

    \item
    \textbf{Pairwise Selection, Sequential Selection and TPA:}
    When pairwise selection, sequential selection and TPA are all active, the cut-off point for sequential selection is based on $\lambda_\text{eff}$.
\end{itemize}

\section{Evolving ES Structures}
\label{sec:ga}

As mentioned in \Cref{sub:es}, the modular framework (\Cref{alg:es}) can be used to instantiate many different ES-structures. There is no interest in the performance of every individual combination, both because there are far too many and because most will perform poorly. Instead, it is far more interesting to determine the best performing ES-structure for different \textit{classes} of optimization problems.

For a GA tasked with this optimization problem, a valid way of comparing ESs is required. The considerations to be taken into account in establishing a robust comparison are examined in \Cref{sub:eval}. Next, \Cref{sub:enc} explains the encoding that is used during the search process. Details of the GA used can be found in \Cref{sub:ga}.

\subsection{Evaluating Evolution Strategies}
\label{sub:eval}

Let $f$ be some (black-box) function to be minimized, i.e. the aim is to approach $\xopt$ as closely as possible, where $\xopt$ is defined by $f(\xopt) \leq f(\bm{x})$ $\forall \bm{x}$. Now, let $\mathbb{O}$ be a set of stochastic continuous optimization methods, e.g. Evolution Strategies (ESs). To keep things simple, the aim is to define a quality measure based on the output of $f(\bm{x})$ to compare any two of these optimization methods.

Let $O \in \mathbb{O}$ be some optimization method of interest. Then we define $y^{O,f} = f(\xopt^O)$\footnote{The function $f$ will be omitted from the notation of $y^{O,f}$ in the following, as comparisons between different functions are meaningless.}, where $\xopt^O$ is the best instance of $\bm{x}$ that was found by a run of optimizer $O$.

Although the optimizers in question are stochastic, a well-tuned optimizer will on average result in much lower values for $\y[]{O}$ than a poorly tuned optimizer. By treating a single $y^O$ as a \textit{sample} from the distribution of possible outcomes, the mean $\avgy{O} = \frac{1}{n} \sum^{n}_{i=1} \y[i]{O}$ of multiple runs can be used as a more stable quality measure because of the Central Limit Theorem, given a large enough sample size $n$.

A value for this parameter $n$ must be chosen as low as possible to reduce computational effort. But, as $n$ increases, the standard error $s^O = \sqrt{\frac{1}{n}\sum_{i=1}^{n} (\y[i]{O} - \avgy{O})^2}$ associated with $\avgy{O}$ will decrease. This can be used to calculate the \textit{uncertainty} of a comparison between any two optimizers.

Let $A, B \in \mathbb{O}$ be two optimizers to be compared.  When $\avgy{A} < \avgy{B}$, we say that $A$ performs better than $B$. However, there is a non-zero probability that $A$ and $B$ are at least indistinguishable in terms of quality. This probability, denoted here as $P(A \equiv B)$, will be used as an uncertainty indication for the comparison $P(A \equiv B) = 2 (1 - cdf(t))$, where $t$ is the Welch's \textit{t-test}
\begin{equation}
    \label{eq:t-test}
    t = \dfrac{\mid \avgy{B} - \avgy{A} \mid}{s_e}
\end{equation}
and the standard error $s_e$ is calculated as $s_e = \sqrt{\frac{(s^A)^2 + (s^B)^2}{n}}$ Furthermore, \textit{cdf} is the cumulative distribution function of the \textit{t-distribution} with $2n-2$ degrees of freedom.

To limit the number of variables for these calculations, the relative distances $d = (\avgy{B} - \avgy{A})/\avgy{A}$ are considered, where $\avgy{A} < \avgy{B}$. The standard errors are proportionally set to $s^B = (1+d)s^A$. Specifically, this reduces \Cref{eq:t-test} to $t = d/s_e$, and establishes the relationship between the parameter $n$ and the uncertainty of comparisons between any two optimizers $A$ and $B$. The experiments that were performed to determine a useful value for this $n$ are described in \Cref{sub:numruns}.

\subsection{Encoding}
\label{sub:enc}

\begin{table}
    \renewcommand{\arraystretch}{1.1}
    \setlength\tabcolsep{5pt}
    \footnotesize
    \caption{\textit{Overview of the available ES modules studied in this paper.} For most of these modules the only required options are \textit{off} and \textit{on}, encoded by the values 0 and 1. For quasi-Gaussian sampling and increasing population, the additional option is encoded by the value 2. The entries in row 9, recombination weights, specify the formula for calculating each weight $w_i$.}
    \label{tab:es-opt}
    \begin{tabular}{lllll}
    \hline
    \# & Module name             & 0 (default)      & 1                 & 2 \\
    \hline
    1  & Active Update           & off              & on                & - \\
    2  & Elitism                 & ($\mu, \lambda$) & ($\mu{+}\lambda$) & - \\
    3  & Mirrored Sampling       & off              & on                & - \\
    4  & Orthogonal Sampling     & off              & on                & - \\
    5  & Sequential Selection    & off              & on                & - \\
    6  & Threshold Convergence   & off              & on                & - \\
    7  & TPA                     & off              & on                & - \\
    8  & Pairwise Selection      & off              & on                & - \\
    9  & Recombination Weights & $\log(\mu{+}\frac{1}{2}){-}\frac{\log(i)}{\sum_j w_j}$ & $\frac{1}{\mu}$& - \\
    10 & Quasi-Gaussian Sampling & off             & Sobol            & Halton \\
    11 & Increasing Population   & off             & IPOP             & BIPOP \\
    \hline
    \end{tabular}
    \vspace{-5pt}
\end{table}

\Cref{tab:es-opt} provides a summary of the ES modules considered in this paper in the same order as introduced in \Cref{sub:es}. By choosing integers to represent the different modules and listing them in the specified order, the structure of an ES can be represented as a list of integers $\vec{r} = r_1 r_2 \ldots r_m$, where $m$ is the number of available module choices. The resulting representations range from $\mathbf{00000000000}$ (default CMA-ES), to $\mathbf{11111111122}$ (CMA-ES with all modules activated).

Decoding a given representation $\vec{r}$ works as follows: For each integer $r_i$ in the representation $\vec{r}$, find the ES module $i$ in \Cref{tab:es-opt}, and use the option indicated by $r_i$. For example, the representation $\vec{r} = \mathbf{01100000100}$ represents the non-default option for ES modules 2, 3 and 9: \textit{elitism}, \textit{mirrored sampling} and \textit{pairwise selection}. In other words: A ($\mu$+$\lambda$)-mirrored-CMA-ES with pairwise selection.

\subsection{Genetic Algorithm}
\label{sub:ga}

A mutation only, self-adaptive GA according to Kruisselbrink \textit{et al.} \cite{kruisselbrink11} is used as optimizer for the ES-structure (see \Cref{alg:ga}). Crossover is omitted to reduce the number of exogenous parameters of the GA. An individual in the GA consists of an ES-structure $\vec{r}$ (previously described in \Cref{sub:enc}) and the self-adaptive mutation rate $p_{\text{m}}$. This algorithm was chosen because of its fast and reliable convergence, as shown in \cite{kruisselbrink11}.

\begin{algorithm}
    \caption{($1, \lambda$)-self-adaptive GA}
    \label{alg:ga}
    \algrenewcommand\algorithmicindent{1.0em}
    \begin{algorithmic}[1]
        \State $t \gets 0$
        \State $P^{(0)} \gets$ generate individual $\vec{I}$, randomly
        \While {not terminate}
            \For {$i=1$ to $\lambda$}  \hspace{30pt} \textit{// Create $\lambda$ offspring}
                \State $(\vec{r}_i, p_{\text{m},i}) = \vec{I}_i \gets$ copy($P^{(t)}$)
                \State $p_{\text{m},i} \gets$ mutate\_$p$($p_{\text{m},i}$) \hspace{0pt} \textit{// Update mutation rate}
                \State $\vec{r}_i \gets$ mutate($\vec{r}_i$, $p_{\text{m},i}$)  \hspace{4pt} \textit{// Update ES structure}
                \State \hspace{99pt}\textit{// with mutation rate $p_{\text{m},i}$}
                \State $f_i \gets$ evaluate($\vec{r}_i$)
            \EndFor
            \State $P^{(t+1)} \gets \vec{I}_{1:\lambda}$, select single best from $\lambda$
            \State $t \gets t + 1$
        \EndWhile
    \end{algorithmic}
\end{algorithm}

The mutation of $\vec{r}$ (line 7) occurs by changing each value $r_j$ in $\vec{r}$ with probability $p_{\text{m},i}$. The value $r_j$ is flipped between zero and one if there are two available options. For the cases where there are three options, a random 50/50 choice is made from the two remaining options, ensuring that a value selected for mutation, is always actually changed.

\section{Experiments}
\label{sec:exp}

The performed experiments and their results are described in this section. The setup for all experiments is listed in \Cref{sub:setup}. Next, the process by which the number of runs per ES was determined is described in \Cref{sub:numruns}, with the final results following in \Cref{sub:results}.

\subsection{Setup}
\label{sub:setup}

All 24 noiseless functions of the well-known \textit{black-box optimization benchmark} (BBOB) suite \cite{hansen:bbob} are used in 2, 3, 5, 10 and 20-D (dimensions), for a total of 120 experiments. As target values are known, \textit{Fixed Cost Error} (FCE) and \textit{Estimated Running Time} (ERT) values can be calculated for every ES. The combination of these two values will be referred to as the \textit{fitness} of an ES. The ERT values are more informative, but can only be calculated if a target value is reached by at least one of the independent runs. Comparisons between ESs are therefore initially done only by comparing ERT values. If an ERT value is only available for one of two ESs, that ES is declared to be better. Only when both ESs do not have an ERT value available, is the FCE value used for the comparison. The BBOB default target value of $10^{-8}$ is used for these experiments.

A (1,12)-GA with a budget of 240 ES-structure evaluations is used, based on initial experiments with various population and budget sizes. Every run of an ES in turn is given a budget of $10^3$D function evaluations. Our framework is written in Python using the \textit{mpi4py} package \cite{mpi4py} and was run on the DAS-4 cluster \cite{das4}, allowing parallelization of both the twelve individuals per generation of the GA and the number of independent runs $n$ (determined in \Cref{sub:numruns}) per ES.

A brute force search over all possible ES-structures with default parameter values from literature is performed to evaluate convergence of the search towards the best ES possible within the framework. Although this brute force search is at the limit of what is computationally practical, it has a large benefit for the GA runs. Instead of running the encoded ES again for every individual the GA has to evaluate, the associated ERT/FCE values can simply be retrieved from storage. This vastly reduces the additional time spent on running the GA and allows us to use the average results from $30$ runs for each experiment.

\subsection{Uncertainty of Comparisons Between ESs}
\label{sub:numruns}

For these initial experiments, a set of $256$ independent runs for around $4\,000$ ES-structures is used. The used ESs are generated by the GA described in \Cref{sub:ga}, spread out over a representative sample of test functions in multiple dimensionalities. When ERT values were available for both ESs, or if only FCE values were available, the distance is easily calculated by $|\avgy{A} - \avgy{B}|$. When only one ERT value was available, the distance is calculated between the FCE of the ES without ERT, and the target value. As the absolute resulting values can differ a lot between different functions and dimensionalities, these distances are calculated only between ES results for each function/dimensionality combination.

First, an empirical distribution of relative distances is obtained from these preliminary runs. For the $40\%$ smallest distances, $d \leq 1$ holds, while $d \geq 100$ is the case for the $30\%$ largest distances. Due to the large spread of these distances, the uncertainty will be calculated for the distances at $5\%$ intervals according to the cumulative distribution.

Next, $100$-fold subsampling is used to simulate having run each ES only $2 \leq n \leq 256$ times. The standard error is calculated for every sample, and averaged over all samples. Small subsamples can be used as accurate representations of expected samples, but for (much) larger samples, this accuracy is lost as the samples always converge to the mean and variance of the set that is sampled from.

\begin{figure}[t]
    \includegraphics[width=.49\textwidth]{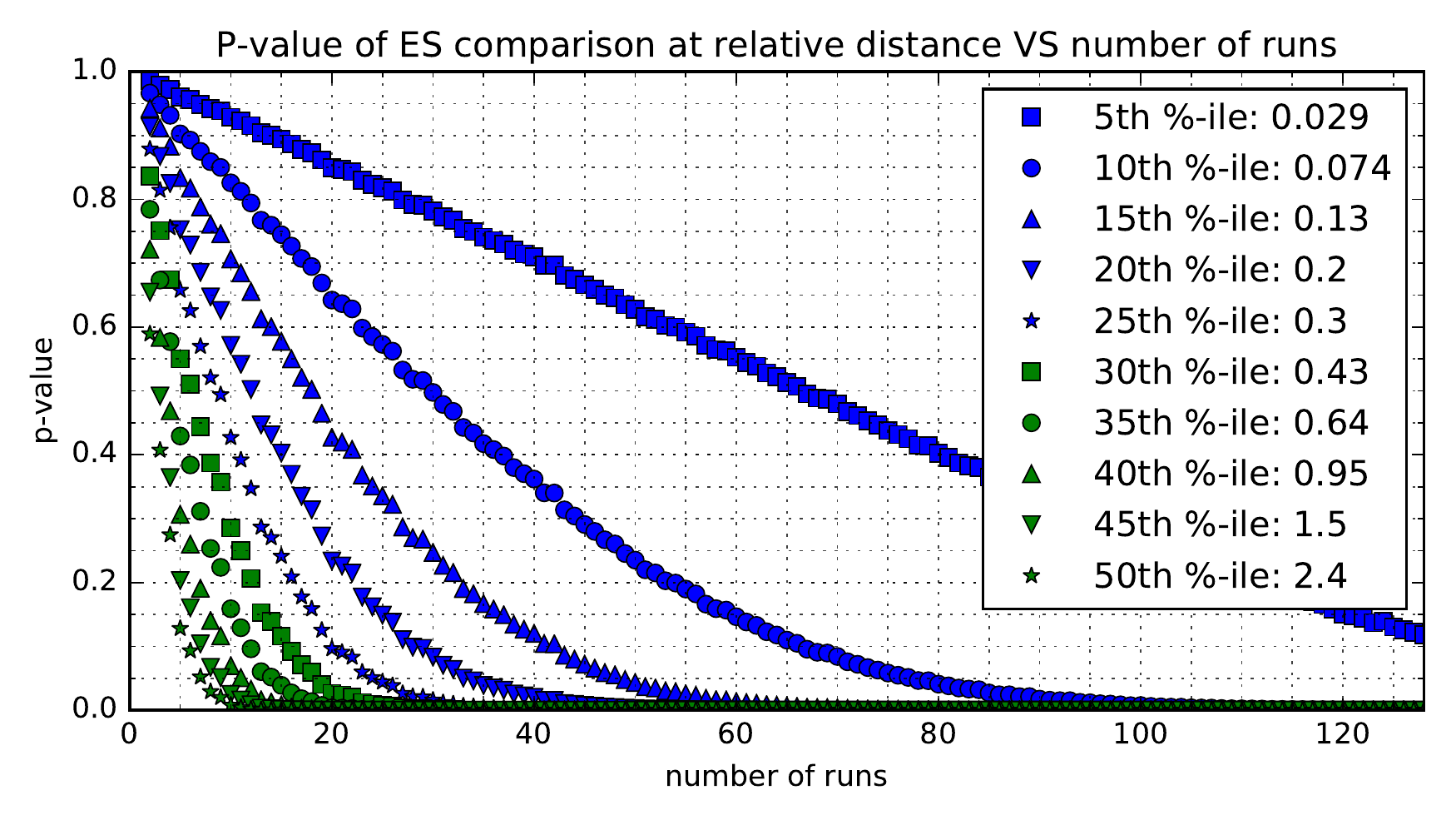}
    \vspace{-15pt}
    \caption{Uncertainty upper bound of comparisons between ESs for the $x$\% smallest distances $d$}
    \label{fig:cert}
    \vspace{-15pt}
\end{figure}

\Cref{fig:cert} shows how the uncertainty decreases with the number of runs. For the 15-20\% smallest distances, there is still a relatively high uncertainty at a large number of runs. As the relative distances are only $20\%$, such high uncertainty is to be expected. For greater relative distances, the uncertainty drops rapidly once a sample size $n \geq 10$ is used.

In this case, we care about the comparison uncertainty because it is of importance to the automated optimization performed by the GA. For all large differences, we want to be very sure that the better one is always chosen, but for increasingly smaller differences, the GA is effectively only building a set of similarly performing ESs. For this reason, we choose to accept $\leq 5$\% uncertainty for the largest $80$\% of all distances as sufficient. Combined with the DAS-4 architecture accommodating $16$ parallel processes per computation node, a value of $n = 32$ is used for all further experiments.

\subsection{Results}
\label{sub:results}

A GA can effectively evolve improving ES structures using our framework. The average runtime of a full brute force run for a single function-dimensionality combination is \textit{between four and five hours} on the DAS-4 cluster. Calculating from this, a single GA run will on average last around 15 minutes.

The evolved ES-structures are often (much) better when comparing the ERT and FCE values of ES-structures found by the GA to those of some standard ES configurations. For example: the ERT values of IPOP-CMA-ES are on average $3.9$ times higher than that of structures found by the GA and at best is only outperformed by $20\%$. Because we are more interested in what kind of structures the GA has produced, further results can be found in \Cref{tab:more_results_1,tab:more_results_2} in the appendix.

\Cref{fig:ga-conv} illustrates that all runs of the GA show convergence in their budget of 20 generations. The behavior of the GA for eight of the ``easier'' functions can be seen in \Cref{subfig:conv-A}. For these functions, most of the possible ES-structures are able to consistently reach the target value of $10^{-8}$, and the GA is able to improve on the ERT over time.

For functions that are more difficult for an ES to optimize, behavior similar to that in \Cref{subfig:conv-B} is seen. Not all ES runs are able to reach the target value, so improvements in the FCE measure can be seen alongside ERT improvements. When none of the runs reach the target value, an ERT value is unavailable. This occurs most often in the 10- and 20-dimensional experiments. Note that the FCE measure is not strictly decreasing when an ERT measure is available, as FCE values are ignored in the comparison between ERT values.

\Cref{subfig:conv-C} shows convergence for functions where the GA only achieves minimal improvements. Whereas the FCE for high-dimensional cases shows improvement in \Cref{subfig:conv-B}, this is much less the case for these functions.

A ranking of the aggregated results by the 30 runs of the GA among all BF runs is shown in \Cref{tab:rank}. This ranking overwhelmingly positive: the aggregated results always rank among the top-20 out of the $4\,608$ possible structures. Although the $0\%$ at rank one indicates that the GA was not able to find the best possible solution in all of the $30$ runs, the $12.5\%$ at rank two shows that the GA often finds the best possible solution.

\begin{table}
    \centering
    \renewcommand{\arraystretch}{1.1}
\caption{\textit{Ranking of GA-found ESs.} The \textit{Fitness} row lists the cumulative percentage out of 120 experiments in which the average ERT/FCE values of the ESs found by the $30$ runs of the GA reaches at least given rank when placed among all results from the BF runs, sorted by fitness.}
\label{tab:rank}
\small
\centering
\begin{tabular}{@{} l|rrrrrrr @{}}
    \toprule
    Rank         & 1   & 2    & 3    & 4--5  & 6--9 & 10--17 & 18+ \\
    Fitness (\%) & 0.0 & 12.5 & 37.5 & 73.33 & 92.5 & 99.17  & 100 \\
    \bottomrule
\end{tabular}
\vspace{-4pt}
\end{table}

\begin{figure*}
    \vspace{-10pt}
    \centering
    \begin{subfigure}[b]{0.32\textwidth}
        \includegraphics[width=\textwidth]{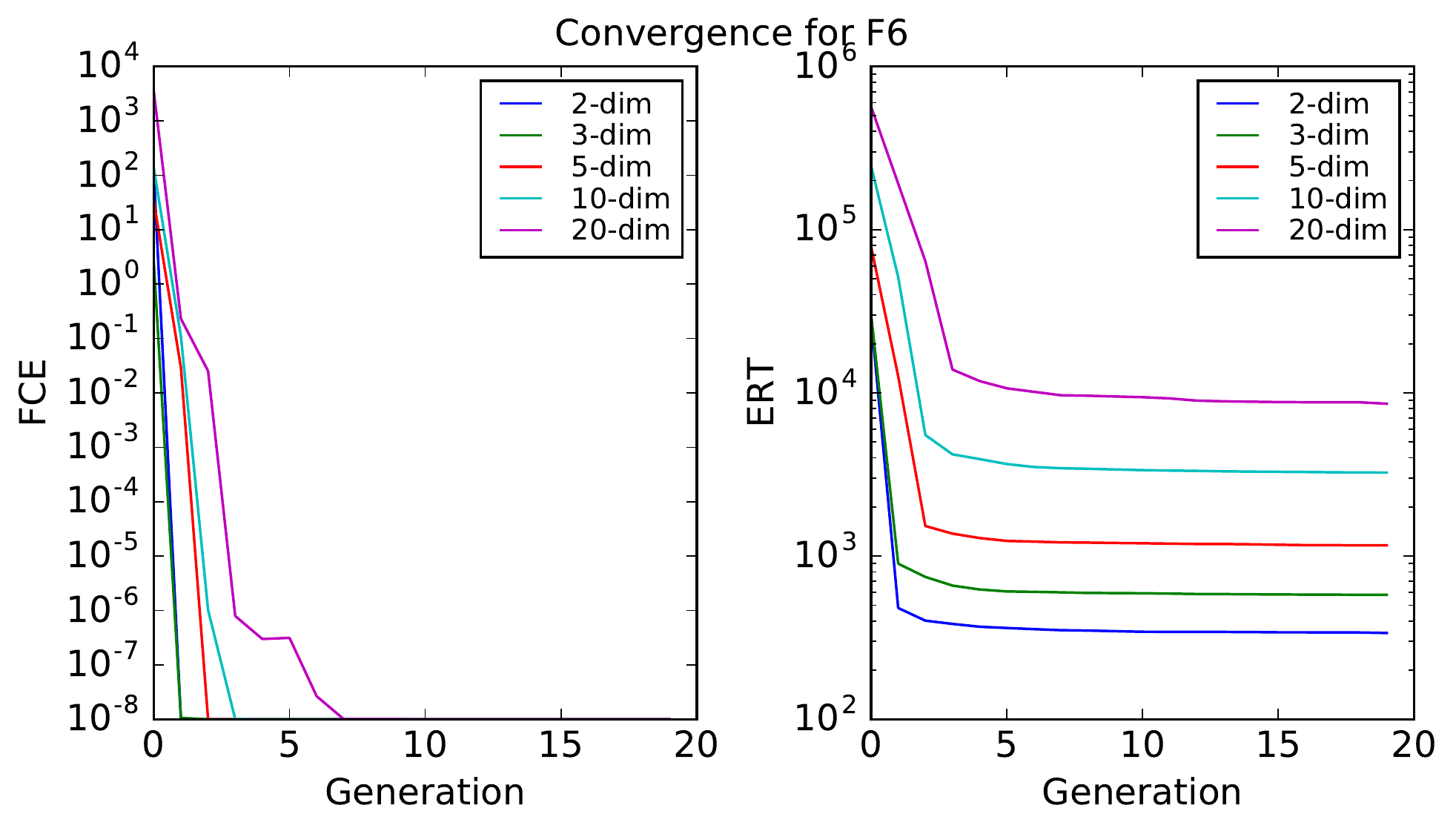}
        \caption{Convergence for F6, similar to 1, 2, 5, 8, 10, 11 and 14.}
        \label{subfig:conv-A}
    \end{subfigure}
    ~
    \begin{subfigure}[b]{0.32\textwidth}
        \includegraphics[width=\textwidth]{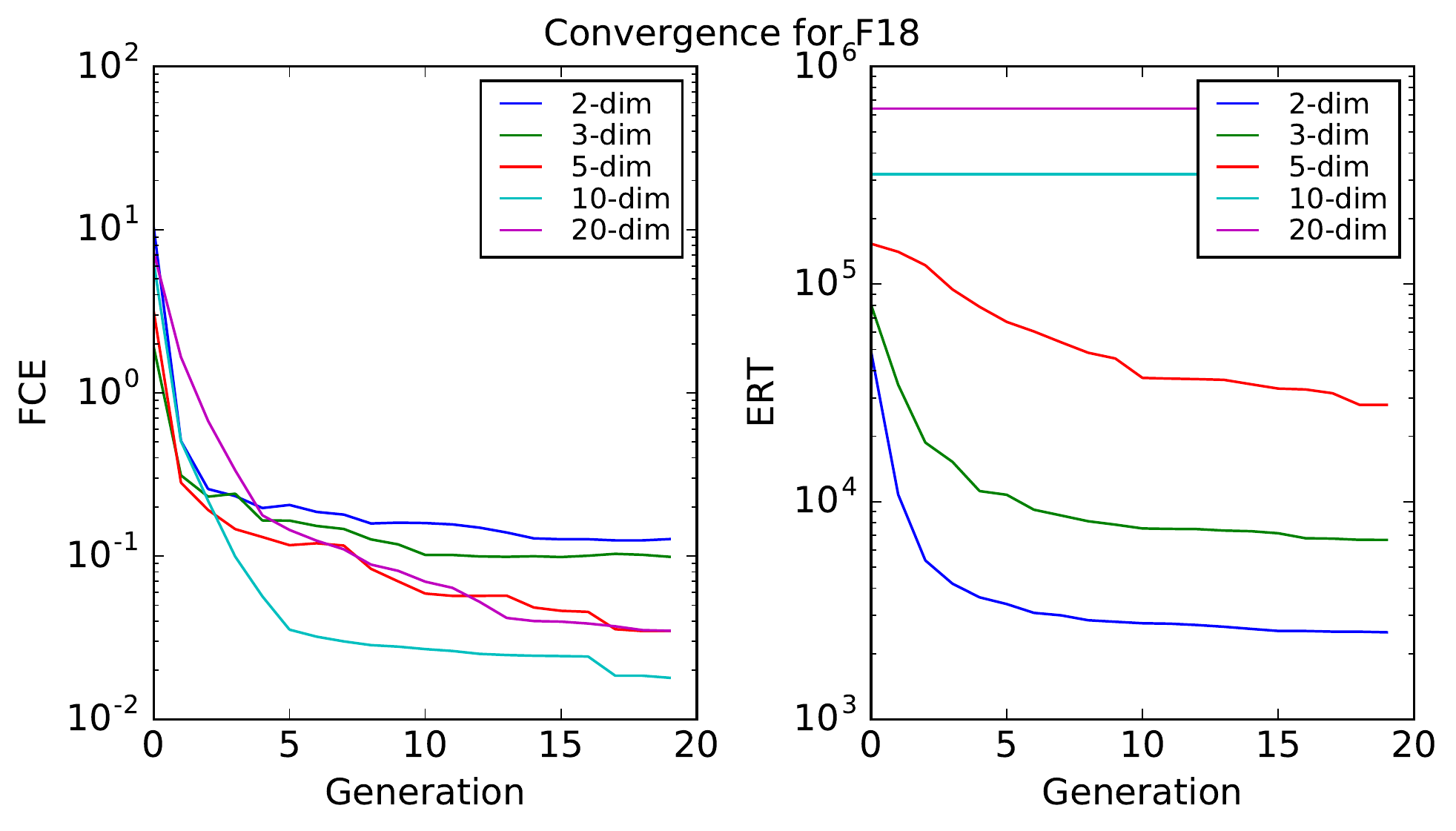}
        \caption{Convergence for F18, similar to 7, 9 13, 16, 17, 23 and 24.}
        \label{subfig:conv-B}
    \end{subfigure}
    ~
    \begin{subfigure}[b]{0.32\textwidth}
        \includegraphics[width=\textwidth]{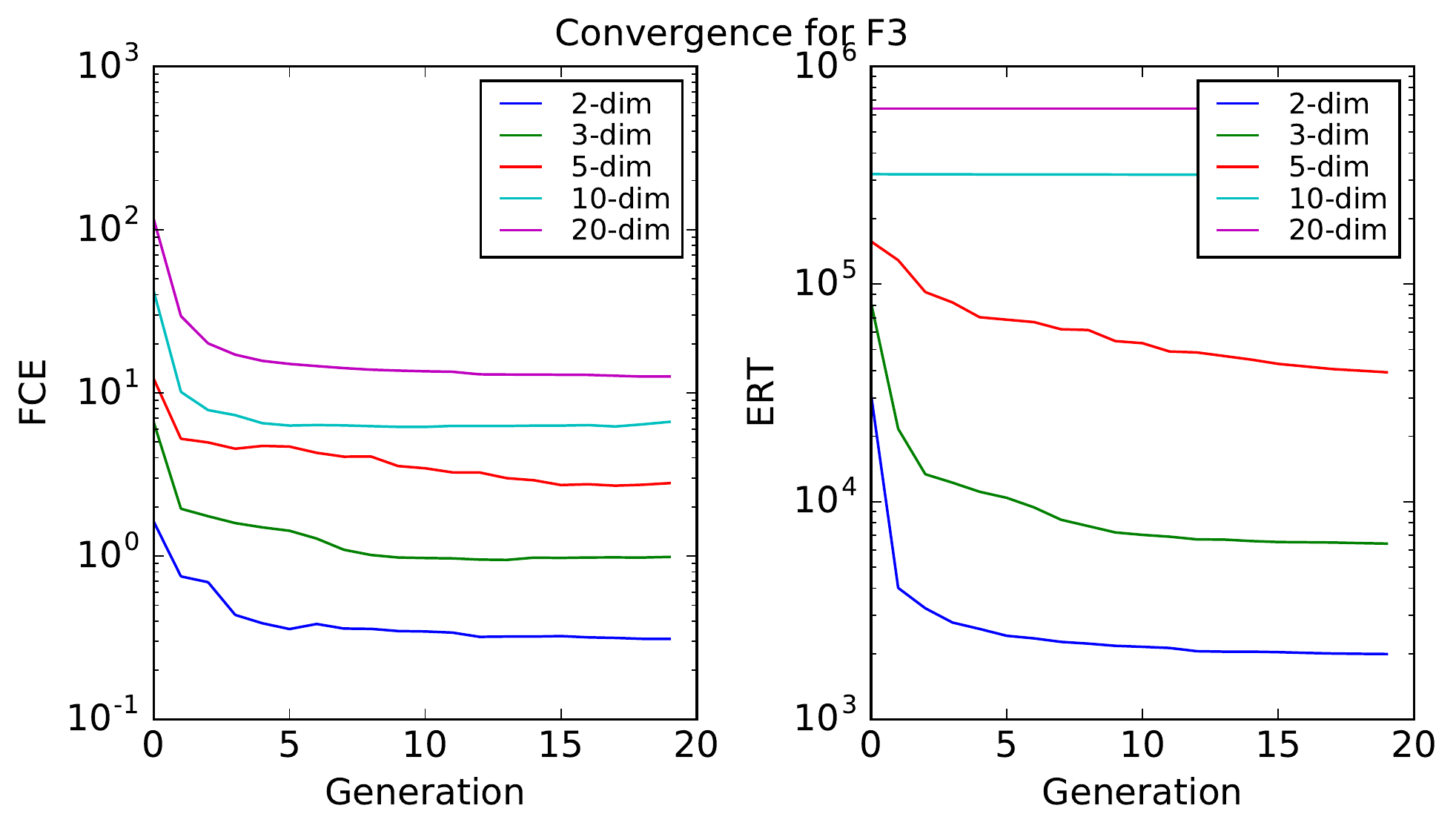}
        \caption{Convergence for F3, similar to 4, 12, 15, 19, 20, 21 and 22.}
        \label{subfig:conv-C}
    \end{subfigure}
    \vspace{-15pt}

    \caption{\textit{GA convergence}. The above graphs show the average rate of convergence during the optimization of ES-structures by $30$ runs of the GA for different BBOB optimization functions. Both the FCE and ERT of the best found structure over time is shown. Results for the optimization process in all five dimensionalities have been plotted per function. As the graphs for groups of several functions are similar, an example has been chosen as representative for each group. The group of functions that a single graph represents is listed below the graph.}
    \label{fig:ga-conv}
    \vspace{-5pt}
\end{figure*}

\begin{table*}
    \centering
\newcommand{\B}[1]{\textbf{#1}}
\newcommand{\BI}[1]{\textbf{\textit{#1}}}
\newcommand{\p}{\phantom{abc}}
\newcommand{\mul}[1]{\multicolumn{1}{c}{#1}}
\newcommand{\mullc}[1]{\multicolumn{2}{c}{#1}}

\renewcommand{\arraystretch}{1.1}
\caption[caption]{\textit{Module activation percentages by function subgroup.} This table lists the frequency of activated algorithmic modules among the experimental cases, obtained by both all $30$ runs per experiment of the Genetic Algorithm (GA) and Brute Force search (BF). Results are separated by the BBOB function subgroups. Except for the second subgroup that consists of four functions, five functions make up each subgroup, for a total of 20 or 25 experiments. For the bottom two rows, the two slash-separated values indicate which of the two choices for the module was chosen. \textbf{Bold} values indicate a relatively \textbf{high} activation percentage for a module in a particular subgroup, while a \textbf{\textit{bold italic}} value indicates a relatively \textbf{\textit{low}} activation percentage.}
\vspace{-5pt}
\label{tab:count}
\small
\begin{tabular}{@{} l *{10}{c} @{}l cc @{}}
	\toprule
	                       & \mullc{F1--F5}& \mullc{F6--F9}  & \mullc{F10--F14}  & \mullc{F15--F19}  &\mullc{F20--F24}&\p & \mullc{Average}       \\
                            \cmidrule(lr){2-3}\cmidrule(lr){4-5}\cmidrule(lr){6-7} \cmidrule(lr){8-9}  \cmidrule(l){10-11}  \cmidrule(lr){13-14}
    Module name            & GA    & BF    & GA     & BF     & GA      & BF      & GA      & BF      & GA    & BF    &    & GA        & BF        \\
    \midrule
	Active Update          & 33    &    36 &     11 &     20 &  \B{40} &  \B{44} &      19 &      20 &    29 &    32 &    & 27.0      &      30.8 \\
	Elitism                & 41    &    44 &     45 &     45 &  \B{70} &  \B{72} & \BI{12} & \BI{12} &    54 &    48 &    & 44.3      &      44.2 \\
	Mirrored Sampling      & 61    &    60 &     62 &     60 & \BI{39} & \BI{40} &      73 &      68 &    59 &    60 &    & 58.4      &      57.5 \\
	Orthogonal Sampling    & 61    &    68 &     54 &     45 &      49 &      44 &      56 &      64 &    51 &    52 &    & 54.1      &      55.0 \\
	Sequential Selection   & 40    &    36 &     39 &     45 &      45 &      48 & \BI{21} & \BI{12} &    30 &    44 &    & 34.8      &      36.7 \\
	Threshold Convergence  & 26    &    36 &     17 &     25 &  \B{53} &  \B{56} &  \BI{2} &  \BI{4} &    15 &    24 &    & 22.6      &      29.2 \\
	TPA                    & 38    &    44 &     35 &     45 &  \B{53} &  \B{56} &  \BI{9} &  \BI{4} &    21 &    28 &    & 31.1      &      35.0 \\
	Pairwise Selection     & 24    &    20 &     23 &     20 &  \BI{6} &  \BI{4} &      20 &      24 &    35 &    20 &    & 21.3      &      17.5 \\
	Recombination Weights  & 16    &    20 & \BI{7} & \BI{5} &  \B{30} &  \B{32} &      20 &      24 &    15 &    12 &    & 17.9      &      19.2 \\
	Sobol/Halton           & 51/32 & 48/32 &  55/32 &  45/45 &   45/47 &   40/56 &   52/35 &   56/40 & 42/38 & 48/44 &    & 48.8/36.8 & 47.5/41.7 \\
	IPOP/BIPOP             & 42/35 & 52/32 &  37/37 &  30/25 &   28/25 &   36/20 &   53/46 &   40/60 & 34/57 & 36/56 &    & 38.9/39.9 & 39.2/39.2 \\
    \bottomrule
\end{tabular}
\end{table*}

\begin{table*}
    \centering
\newcommand{\B}[1]{\textbf{#1}}
\newcommand{\BI}[1]{\textbf{\textit{#1}}}
\newcommand{\p}{\phantom{abc}}
\newcommand{\mul}[1]{\multicolumn{1}{c}{#1}}
\newcommand{\mullc}[1]{\multicolumn{2}{c}{#1}}

\renewcommand{\arraystretch}{1.1}
\caption{\textit{Module activation percentages by dimensionality.} This table lists the frequency of activated algorithmic modules among the experimental cases, obtained by both all $30$ runs per experiment of the Genetic Algorithm (GA) and Brute Force search (BF). For the bottom two rows, the two slash-separated values indicate which of the two choices for the module was chosen. \textbf{Bold} values indicate a relatively \textbf{high} activation percentage for a module in a particular subgroup, while a \textbf{\textit{bold italic}} value indicates a relatively \textbf{\textit{low}} activation percentage.}
\vspace{-5pt}
\label{tab:count_dim}
\small
\begin{tabular}{@{} l *{10}{c} @{}l cc @{}}
	\toprule
	                       & \mullc{2D    }& \mullc{3D}    & \mullc{5D}    & \mullc{10D}   &\mullc{20D}    &\p & \mullc{Average}        \\
                            \cmidrule(lr){2-3}\cmidrule(lr){4-5}\cmidrule(lr){6-7} \cmidrule(lr){8-9}  \cmidrule(l){10-11}  \cmidrule(lr){13-14}
    Module name            & GA    & BF    & GA    & BF    & GA    & BF    & GA    & BF    & GA    & BF    &    & GA        & BF        \\
    \midrule
	Active Update          & \B{53}& \B{67}& 40    & 50    & 25    & 17    &  5    &  8    & 13    & 13    &    & 27.0      &      30.8 \\
	Elitism                & \B{72}& \B{75}& \B{66}& \B{71}& 42    & 46    & 23    & 13    & 19    & 17    &    & 44.3      &      44.2 \\
	Mirrored Sampling      &\BI{36}&\BI{33}& 50    & 42    & 66    & 71    & \B{72}& \B{75}& \B{68}& \B{67}&    & 58.4      &      57.5 \\
	Orthogonal Sampling    & 38    & 37    & 35    & 33    & 60    & 75    & \B{74}& \B{75}& 63    & 54    &    & 54.1      &      55.0 \\
	Sequential Selection   & 47    & 46    & 42    & 46    & 29    & 25    & 31    & 33    & 24    & 33    &    & 34.8      &      36.7 \\
	Threshold Convergence  & 35    & 42    & 38    & 50    & 19    & 29    &\BI{11}&\BI{13}&\BI{10}&\BI{13}&    & 22.6      &      29.2 \\
	TPA                    & 41    & 46    & 45    & 50    & 30    & 42    & 22    & 21    & 18    & 17    &    & 31.1      &      35.0 \\
	Pairwise Selection     & 21    & 21    & 27    & 25    & 18    &  8    & 16    &  8    & 24    & 25    &    & 21.3      &      17.5 \\
	Recombination Weights  & 24    & 25    & 16    & 21    & 18    & 25    & 18    & 17    & 14    &  8    &    & 17.9      &      19.2 \\
	Sobol/Halton           & 47/43 & 33/54 & 51/38 & 54/42 & 27/63 & 21/71 & 55/26 & 67/25 & 63/14 & 63/17 &    & 48.8/36.8 & 47.5/41.7 \\
	IPOP/BIPOP             & 35/39 & 38/42 & 34/44 & 25/46 & 38/39 & 46/33 & 37/51 & 38/54 & 50/27 & 50/21 &    & 38.9/39.9 & 39.2/39.2 \\
    \bottomrule
\end{tabular}
\vspace{-5pt}

\end{table*}

Having established that most GA-found results correspond to the best ES-structures found by BF, \Cref{tab:count} shows the relative activation of the eleven available modules over all experiments. These results again show a lot of similarity between the GA and BF runs, with minor differences. Overall, some modules are only activated in few cases such as alternative recombination weights ($15$--$20\%$) and threshold convergence ($23$--$29\%$), while others seem to be activated more evenly such as Orthogonal Sampling and Elitism with total activation percentages around $50\%$. Most successful are Mirrored Sampling ($58\%$), Increasing Population ($78\%$) and Quasi-Gaussian Sampling ($84\%$).

Separating the activation percentages by function subgroups provides some additional insight. For most subgroups, the percentages correspond quite well to the aggregated values, but the third and fourth subgroups (F10--F14 and F15--F20) show interesting behavior. \Cref{tab:count} illustrates this with highlighted outlying values. All higher than average activation percentages occur with the third subgroup of unimodal functions with high conditioning. Especially Elitism with Threshold Convergence seems to be very effective for this group of functions. Meanwhile, most of the lower than average percentages are for the fourth subgroup of multi-modal functions with adequate global structure. Elitism, Threshold Convergence and Two-Point step-size Adaptation are almost never active in the best performing ESs for these functions. This is almost a complete opposite of the results for the third subgroup.

\Cref{tab:count_dim} separates the activation percentages by problem dimensionality. This highlights a trend of increasing or decreasing performance for some of the available modules. Active, Elitism and Threshold Convergence for example are selected more often for lower dimensionality problems, while Mirrored and Orthogonal Sampling are increasingly activated for problems with higher dimensionality. Combined with the consistently high activation percentages of Quasi-Gaussian sampling, these results suggest that especially in high-dimensional problems, the success of an ES depends on its ability to uniformly sample the neighboring search space.

A particular case to examine is the correlation between Mirrored Sampling and Pairwise Selection, because of their paired introduction to reduce sampling bias. With exception of the third subgroup, Mirrored Sampling is active in $60\%$ or more of all cases. Pairwise selection however is only active in $20$--$25\%$ of the experiments, or only one third as often. A similar effect is seen also in the third subgroup, where both modules are only activated half as often as in the other subgroups, maintaining the ratio between them.

\section{Conclusions and Outlook}
\label{sec:conc}

By extracting structural features from different CMA-ES variations, a modular framework for running new ES-structures is created, in which the structures (configurations) of CMA-ES are optimized for different optimization problems. ES-modules considered in this work are implemented in a way that they can be activated independently, with minimal dependency checking. In addition, this framework is also extensible for other modules that are not considered in this research.

Based on the empirical study using the well-known BBOB function suite, is it clear that the proposed approach on ES structure configuration, exploiting a self-adaptive genetic algorithm, can consistently produce results that are comparable to the top $0.5\%$ of the best results from the brute force search. The GA achieves such a result on $240$ ES-structures, constituting only $5\%$ of the whole configuration space, showing the advantage of the GA search when more ES structure modules are incorporated into the framework. Furthermore, note that in real applications where function evaluations are costly, the overhead on determining the fitness of ES structures could grow drastically. In this case, the proposed GA-based search is more computational tractable than the exhaustive search.

Across function groups and problem dimensions, the analysis of the selected ES-structures clearly confirms that different modules or ES-variants may excel in some cases, at the cost of reduced performance for other cases. Overall, the most successful modules are Increasing Population and Quasi-Gaussian Sampling. Note that (B)IPOP is chosen very often, even with the limited budget of $10^3$D function evaluations. Elitism, Two-Point Step-Size Adaptation and Threshold Convergence perform best for unimodal functions with high conditioning. Alternatively, Elitism, Active update and Sequential Selection are recommended for low dimensionality problems, while for increasingly higher dimensionalities, use of Mirrored and Orthogonal Sampling give the most likely increase in performance.

In the future research, the impact of each ES module can be analyzed by constructing a data-driven model (e.g. decision tree) on the ES structures and their performance measures. The most influential ES module on a particular function class can be related to the landscape features (e.g. convexity, ruggedness) of this function. Such relations could help to verify theoretical hypothesis on the optimization problem or even gain more knowledge on the problem. The proposed framework should be extended with other modules, which make the configuration space even larger. In addition, this framework should be tested on some problem class that are summarized from complex real-world optimization problems.

\section{Acknowledgments}
The authors would like to thank the reviewers for their helpful comments.

\bibliographystyle{abbrv}
\bibliography{svrijn}

\appendix
\label{app:commons}

\begin{table}[h]
    \footnotesize
    \centering
    \caption{\textit{Common ES Variants} A selection of ten common ES variants is listed here, as referred to by \Cref{tab:more_results_1,tab:more_results_2}.}
    \label{tab:commons}
    \begin{tabular}{@{} ll @{}}
        \toprule
        Variant                     & Representation \\
        \midrule
        CMA-ES                      & 00000000000    \\
        Active CMA-ES               & 10000000000    \\
        Elitist CMA-ES              & 01000000000    \\
        Mirrored-pairwise CMA-ES    & 00100001000    \\
        IPOP-CMA-ES                 & 00000000001    \\
        Active IPOP-CMA-ES          & 10000000001    \\
        Elitist Active IPOP-CMA-ES  & 11000000001    \\
        BIPOP-CMA-ES                & 00000000002    \\
        Active BIPOP-CMA-ES         & 10000000002    \\
        Elitist Active BIPOP-CMA-ES & 11000000002    \\
        \bottomrule
    \end{tabular}
\end{table}

\begin{table*}[h]
    \centering
    \small
\caption{\textit{Comparison of GA-found structures with common ES variants for F1--F12} For each function-dimensionality combination, the best ES-structure as evolved by any of the $30$ runs of the GA is compared to the best out of a standard set of common ES variants (see \cref{tab:commons}). All results have been taken from the brute force search. The ``Relative'' column shows either the ERTs or FCEs of the best common variant divided by that of the structure found by the GA. Values greater than $1$ indicate that the GA solution is better, values under $1$ indicate that the best common variant is better. A relative value is in \textit{italics} when an ERT value is only available for the GA-found solution, but the FCE value for the best common variant is better.}
\label{tab:more_results_1}
\centering
\begin{tabular}{@{} crcrrcrrr @{}}
    \toprule
    F-ID & N & Best Common Variant & ERT & FCE & Best found by GA & ERT & FCE & Relative \\
    \midrule
    F1  &  2 & 11000000001 &    294.75 &     1e-08 & 11110111000 &    165.25 &     1e-08 &     1.78 \\
    F1  &  3 & 11000000001 &   473.594 &     1e-08 & 11110111010 &   235.375 &     1e-08 &     2.01 \\
    F1  &  5 & 11000000001 &     797.5 &     1e-08 & 10110010021 &       440 &     1e-08 &     1.81 \\
    F1  & 10 & 01000000000 &   1634.38 &     1e-08 & 00110110022 &   726.438 &     1e-08 &     2.25 \\
    F1  & 20 & 11000000001 &   3050.25 &     1e-08 & 00110110020 &   1285.38 &     1e-08 &     2.37 \\
    \hline
    F2  &  2 & 11000000001 &   660.375 &     1e-08 & 11011110021 &   434.062 &     1e-08 &     1.52 \\
    F2  &  3 & 01000000000 &   1160.03 &     1e-08 & 01001110012 &   678.594 &     1e-08 &     1.71 \\
    F2  &  5 & 00000000002 &   2314.75 &     1e-08 & 11000110022 &    1364.5 &     1e-08 &      1.7 \\
    F2  & 10 & 00000000002 &   6428.44 &     1e-08 & 00011000022 &   5132.06 &     1e-08 &     1.25 \\
    F2  & 20 & 00000000002 &   45225.4 & 0.0001362 & 00001000011 &   16757.1 &     1e-08 &      2.7 \\
    \hline
    F3  &  2 & 10000000001 &   2742.35 &    0.3582 & 10111000012 &   1795.73 &    0.2912 &     1.53 \\
    F3  &  3 & 10000000002 &     12346 &     1.062 & 00100010011 &    5380.4 &    0.8374 &     2.29 \\
    F3  &  5 & 11000000002 &     76816 &     5.074 & 10111000021 &   29318.8 &     2.067 &     2.62 \\
    F3  & 10 & 00000000002 &       N/A &      7.12 & 00001000011 &    314990 &     6.102 &     1.17 \\
    F3  & 20 & 00000000001 &       N/A &     17.63 & 00110000011 &       N/A &      12.3 &     1.43 \\
    \hline
    F4  &  2 & 11000000002 &   11877.6 &     1.464 & 11100100002 &      4089 &    0.8781 &      2.9 \\
    F4  &  3 & 11000000002 &     93793 &     4.469 & 11010001112 &   17803.2 &     3.561 &     5.27 \\
    F4  &  5 & 00000000001 &       N/A &     5.905 & 01110111121 &    155980 &     8.998 &\textit{0.656}\\
    F4  & 10 & 00000000002 &       N/A &     13.12 & 00110000011 &       N/A &     9.929 &     1.32 \\
    F4  & 20 & 00000000002 &       N/A &     31.03 & 00100000011 &       N/A &     24.51 &     1.27 \\
    \hline
    F5  &  2 & 11000000001 &   605.812 &     1e-08 & 01011000121 &   466.594 &     1e-08 &      1.3 \\
    F5  &  3 & 00100001000 &   940.688 &     1e-08 & 01111000100 &   736.656 &     1e-08 &     1.28 \\
    F5  &  5 & 00000000001 &      1731 &     1e-08 & 01011000011 &   1324.28 &     1e-08 &     1.31 \\
    F5  & 10 & 00100001000 &   3505.31 &     1e-08 & 00000000012 &   3054.69 &     1e-08 &     1.15 \\
    F5  & 20 & 00000000001 &     10464 &    0.3436 & 10110001001 &   7143.75 &     1e-08 &     1.46 \\
    \hline
    F6  &  2 & 01000000000 &   428.438 &     1e-08 & 01000011020 &   315.062 &     1e-08 &     1.36 \\
    F6  &  3 & 01000000000 &       805 &     1e-08 & 01110001020 &   573.375 &     1e-08 &      1.4 \\
    F6  &  5 & 01000000000 &      1574 &     1e-08 & 01111000021 &   1141.78 &     1e-08 &     1.38 \\
    F6  & 10 & 00100001000 &   4114.69 &     1e-08 & 00110000012 &   3151.56 &     1e-08 &     1.31 \\
    F6  & 20 & 00100001000 &   9618.38 &     1e-08 & 00110001011 &      8385 &     1e-08 &     1.15 \\
    \hline
    F7  &  2 & 00000000001 &    438.75 &     1e-08 & 11110011022 &       266 &     1e-08 &     1.65 \\
    F7  &  3 & 00000000001 &   886.516 &  0.002689 & 01000110022 &   540.719 &     1e-08 &     1.64 \\
    F7  &  5 & 10000000002 &    2587.5 &     1e-08 & 00000010021 &   1613.56 &     1e-08 &      1.6 \\
    F7  & 10 & 00000000001 &    9176.4 &   0.04231 & 00000010011 &   5894.97 &  0.007247 &     1.56 \\
    F7  & 20 & 00000000001 &    153270 &     1.093 & 00000110011 &   21123.9 &    0.0704 &     7.26 \\
    \hline
    F8  &  2 & 11000000001 &   574.125 &     1e-08 & 11000110010 &   422.312 &     1e-08 &     1.36 \\
    F8  &  3 & 01000000000 &   1034.69 &     1e-08 & 01001000120 &   827.438 &     1e-08 &     1.25 \\
    F8  &  5 & 01000000000 &   2705.43 &    0.4914 & 00111000021 &   1705.62 &     1e-08 &     1.59 \\
    F8  & 10 & 00100001000 &   6843.87 &    0.1246 & 00111000002 &   5168.25 &     1e-08 &     1.32 \\
    F8  & 20 & 00100001000 &    105992 &    0.5413 & 00101000010 &   19080.7 & 1.954e-07 &     5.55 \\
    \hline
    F9  &  2 & 01000000000 &   546.938 &     1e-08 & 11000110020 &   423.062 &     1e-08 &     1.29 \\
    F9  &  3 & 00000000000 &   1183.66 &     1e-08 & 11101110010 &    764.25 &     1e-08 &     1.55 \\
    F9  &  5 & 00000000002 &   2607.23 &  0.000213 & 00111000000 &   1710.22 &     1e-08 &     1.52 \\
    F9  & 10 & 00000000001 &      6635 &     1e-08 & 00111000012 &   5187.56 &     1e-08 &     1.28 \\
    F9  & 20 & 00100001000 &   48317.5 &    0.2955 & 00101000010 &     18609 & 2.243e-07 &      2.6 \\
    \hline
    F10 &  2 & 11000000001 &   662.062 &     1e-08 & 11001110020 &   441.688 &     1e-08 &      1.5 \\
    F10 &  3 & 11000000001 &   1146.25 &     1e-08 & 11001110012 &     693.5 &     1e-08 &     1.65 \\
    F10 &  5 & 01000000000 &      2228 &     1e-08 & 01000110021 &   1458.25 &     1e-08 &     1.53 \\
    F10 & 10 & 00000000000 &   6509.06 &     1e-08 & 00111000012 &    5181.5 &     1e-08 &     1.26 \\
    F10 & 20 & 00000000000 &   35179.3 & 2.703e-06 & 00001000010 &   16564.5 &     1e-08 &     2.12 \\
    \hline
    F11 &  2 & 11000000001 &   629.625 &     1e-08 & 11011110021 &   453.188 &     1e-08 &     1.39 \\
    F11 &  3 & 00000000001 &   1334.38 &     1e-08 & 11110110010 &   734.969 &     1e-08 &     1.82 \\
    F11 &  5 & 00000000001 &      2493 &     1e-08 & 11110110120 &   1518.75 &     1e-08 &     1.64 \\
    F11 & 10 & 00000000002 &   6163.12 &     1e-08 & 01110110121 &   3093.31 &     1e-08 &     1.99 \\
    F11 & 20 & 00000000000 &   16148.2 &     1e-08 & 00010110021 &    4901.5 &     1e-08 &     3.29 \\
    \hline
    F12 &  2 & 10000000000 &   1360.85 &    0.1986 & 01101000111 &   776.067 &     1e-08 &     1.75 \\
    F12 &  3 & 01000000000 &    3426.5 &     390.5 & 11000110011 &   2084.69 &   0.02808 &     1.64 \\
    F12 &  5 & 01000000000 &   6489.26 &    0.1195 & 00110000010 &   4017.19 &  1.03e-05 &     1.62 \\
    F12 & 10 & 00000000002 &   33717.8 &  0.003169 & 00111000010 &     12018 & 0.0005758 &     2.81 \\
    F12 & 20 & 00100001000 &     68624 &   0.02719 & 01001000010 &   33516.2 &  0.002301 &     2.05 \\
    \bottomrule
\end{tabular}
\end{table*}

\begin{table*}[h]
    \centering
    \small
\caption{\textit{Comparison of GA-found structures with common ES variants for F13--F24} For each function-dimensionality combination, the best ES-structure as evolved by any of the $30$ runs of the GA is compared to the best out of a standard set of common ES variants (see \cref{tab:commons}). All results have been taken from the brute force search. The ``Relative'' column shows either the ERTs or FCEs of the best common variant divided by that of the structure found by the GA. Values greater than $1$ indicate that the GA solution is better, values under $1$ indicate that the best common variant is better. A relative value is in \textit{italics} when an ERT value is only available for the GA-found solution, but the FCE value for the best common variant is better.}
\label{tab:more_results_2}
\centering
\begin{tabular}{@{} crcrrcrrr @{}}
    \toprule
    F-ID & N & Best Common Variant & ERT & FCE & Best found by GA & ERT & FCE & Relative \\
    \midrule
    F13 &  2 & 10000000000 &   971.812 &     1e-08 & 11000110021 &     518.5 &     1e-08 &     1.87 \\
    F13 &  3 & 10000000001 &   1704.28 &     1e-08 & 11001110020 &   883.344 &     1e-08 &     1.93 \\
    F13 &  5 & 10000000000 &   3830.93 & 3.862e-08 & 01100000122 &   2967.23 &     1e-08 &     1.29 \\
    F13 & 10 & 00000000001 &   44828.6 &  0.007853 & 00010000021 &   13653.3 & 0.0008427 &     3.28 \\
    F13 & 20 & 00000000000 &       N/A &    0.3405 & 10110001000 &    157122 &    0.2778 &     1.23 \\
    \hline
    F14 &  2 & 11000000002 &   618.938 &     1e-08 & 11001110122 &   410.781 &     1e-08 &     1.51 \\
    F14 &  3 & 11000000002 &   1196.12 &     1e-08 & 11001110021 &    630.75 &     1e-08 &      1.9 \\
    F14 &  5 & 01000000000 &   2506.75 &     1e-08 & 01000110120 &      1327 &     1e-08 &     1.89 \\
    F14 & 10 & 01000000000 &      7180 &     1e-08 & 01101000112 &   5056.84 &     1e-08 &     1.42 \\
    F14 & 20 & 01000000000 &    639252 & 1.518e-07 & 01010000120 &   20453.4 &     1e-08 &     31.3 \\
    \hline
    F15 &  2 & 10000000001 &   2864.47 &    0.5562 & 10001000022 &   1881.09 &    0.2853 &     1.52 \\
    F15 &  3 & 00000000001 &      7889 &      1.29 & 00100000012 &   3882.94 &    0.6265 &     2.03 \\
    F15 &  5 & 10000000002 &     51880 &     2.363 & 00110000022 &     17811 &     1.492 &     2.91 \\
    F15 & 10 & 00000000001 &       N/A &     5.769 & 10110000012 &    311980 &     6.078 &\textit{0.949}\\
    F15 & 20 & 00000000002 &       N/A &     15.12 & 00110000011 &       N/A &     8.017 &     1.89 \\
    \hline
    F16 &  2 & 00000000001 &     897.8 &  0.003917 & 00111000011 &   740.367 &  0.001413 &     1.21 \\
    F16 &  3 & 00000000002 &   3739.84 &   0.06716 & 00000000012 &   1490.07 &   0.03927 &     2.51 \\
    F16 &  5 & 00000000001 &   16217.8 &   0.07463 & 00110001021 &   4970.18 &    0.0417 &     3.26 \\
    F16 & 10 & 00000000001 &    105890 &    0.1674 & 00010000011 &     78905 &    0.0878 &     1.34 \\
    F16 & 20 & 00000000001 &       N/A &    0.3336 & 00100001012 &       N/A &    0.1026 &     3.25 \\
    \hline
    F17 &  2 & 11000000002 &   1643.77 & 0.0001774 & 11000000111 &       952 &  9.01e-07 &     1.73 \\
    F17 &  3 & 11000000002 &   5332.69 &   0.01492 & 00001000021 &   1610.94 & 2.252e-07 &     3.31 \\
    F17 &  5 & 00000000001 &     79964 &   0.01337 & 00110000012 &   4393.66 & 5.068e-07 &     18.2 \\
    F17 & 10 & 00000000001 &       N/A &   0.01218 & 10110000002 &     29918 &  0.002855 &     4.27 \\
    F17 & 20 & 00000000001 &       N/A &   0.01415 & 00110000011 &    312924 & 0.0008419 &     16.8 \\
    \hline
    F18 &  2 & 11000000002 &      7734 &    0.3329 & 01000000122 &      2210 &    0.1135 &      3.5 \\
    F18 &  3 & 00100001000 &   22480.5 &    0.1571 & 11000111012 &    4807.5 &    0.1481 &     4.68 \\
    F18 &  5 & 00000000002 &       N/A &    0.1211 & 00110000022 &   22228.6 &    0.0141 &     8.59 \\
    F18 & 10 & 00000000002 &       N/A &   0.07986 & 00110001011 &       N/A &   0.01037 &      7.7 \\
    F18 & 20 & 00000000002 &       N/A &    0.1465 & 00110001002 &       N/A &   0.02336 &     6.27 \\
    \hline
    F19 &  2 & 11000000002 &   3498.86 &   0.01915 & 00010000012 &   1833.39 &  0.004905 &     1.91 \\
    F19 &  3 & 10000000001 &   22671.2 &    0.2385 & 00110001121 &     12702 &    0.2473 &     1.78 \\
    F19 &  5 & 10000000000 &       N/A &     0.405 & 00100000121 &     77184 &    0.1947 &     2.08 \\
    F19 & 10 & 10000000002 &       N/A &     0.666 & 00110000112 &       N/A &    0.2509 &     2.65 \\
    F19 & 20 & 10000000002 &       N/A &     0.801 & 00110000102 &       N/A &    0.3586 &     2.23 \\
    \hline
    F20 &  2 & 11000000001 &    3022.4 &    0.2513 & 11101000012 &   1571.65 &    0.2311 &     1.92 \\
    F20 &  3 & 11000000002 &   29992.7 &    0.8192 & 11111110012 &   7379.73 &    0.5696 &     4.06 \\
    F20 &  5 & 10000000001 &    156448 &     1.313 & 11010110022 &   50956.7 &     0.924 &     3.07 \\
    F20 & 10 & 00000000002 &       N/A &     1.552 & 00110111120 &       N/A &     1.305 &     1.19 \\
    F20 & 20 & 00000000001 &       N/A &     1.634 & 11011001022 &       N/A &     1.519 &     1.08 \\
    \hline
    F21 &  2 & 11000000002 &    459.75 &     1e-08 & 11111000022 &   352.969 &     1e-08 &      1.3 \\
    F21 &  3 & 11000000001 &   2074.62 &    0.1948 & 11101000012 &   1265.62 &   0.09906 &     1.64 \\
    F21 &  5 & 11000000001 &    6536.5 &    0.9468 & 11110000022 &    4443.6 &    0.6978 &     1.47 \\
    F21 & 10 & 00000000001 &   12228.2 &     1.525 & 00110000011 &      8900 &    0.7913 &     1.37 \\
    F21 & 20 & 11000000001 &     42659 &     2.036 & 01111000012 &   18566.9 &    0.8026 &      2.3 \\
    \hline
    F22 &  2 & 11000000002 &   431.625 &     1e-08 & 11000001012 &       360 &     1e-08 &      1.2 \\
    F22 &  3 & 11000000002 &   1609.25 &   0.08648 & 01000000022 &   1136.71 &   0.05941 &     1.42 \\
    F22 &  5 & 11000000002 &   5198.74 &     1.097 & 01111000022 &   3573.12 &    0.2994 &     1.45 \\
    F22 & 10 & 00000000001 &    101163 &     2.357 & 01101000012 &   18711.5 &    0.8101 &     5.41 \\
    F22 & 20 & 00000000002 &    317730 &     3.805 & 00100010001 &    116939 &     4.153 &     2.72 \\
    \hline
    F23 &  2 & 00000000002 &      2652 &    0.2817 & 00000000021 &    1528.8 &    0.1039 &     1.73 \\
    F23 &  3 & 00000000002 &   7371.58 &    0.4927 & 00000001021 &   3794.21 &    0.2224 &     1.94 \\
    F23 &  5 & 00000000000 &   52277.3 &    0.5859 & 00010000020 &   21265.1 &    0.2673 &     2.46 \\
    F23 & 10 & 10000000002 &       N/A &    0.8297 & 00110000022 &       N/A &   0.09923 &     8.36 \\
    F23 & 20 & 10000000002 &       N/A &    0.9746 & 00100000011 &       N/A &   0.05987 &     16.3 \\
    \hline
    F24 &  2 & 00000000002 &     63342 &      2.09 & 00000111101 &   4411.54 &     1.501 &     14.4 \\
    F24 &  3 & 00000000001 &       N/A &     4.007 & 10011100122 &     30994 &     6.622 &\textit{0.605}\\
    F24 &  5 & 10000000002 &       N/A &     7.855 & 00111000012 &       N/A &     5.931 &     1.32 \\
    F24 & 10 & 10000000000 &       N/A &     19.17 & 00100010011 &       N/A &     12.26 &     1.56 \\
    F24 & 20 & 10000000002 &       N/A &     41.37 & 00001000011 &       N/A &     30.08 &     1.38 \\
    \bottomrule
\end{tabular}
\end{table*}

\end{document}